%% file: main.tex
\documentclass[10pt,twocolumn]{article}
\usepackage[a4paper,margin=0.72in,columnsep=0.25in]{geometry}
\usepackage[T1]{fontenc}
\usepackage[utf8]{inputenc}
\usepackage{lmodern}
\usepackage{microtype}
\usepackage{amsmath,amssymb,bm}
\usepackage{graphicx,booktabs,array,multirow}
\usepackage[numbers,sort&compress]{natbib}
\usepackage[font=small,labelfont=bf]{caption}
\usepackage{xcolor}
\usepackage{placeins}
\usepackage{float}
\usepackage{titlesec}
\titleformat{\section}{\large\bfseries\raggedright}{\thesection}{.6em}{}
\titleformat{\subsection}{\normalsize\bfseries\raggedright}{\thesubsection}{.6em}{}
\usepackage{authblk}
\usepackage{balance}
\usepackage{hyperref}
\hypersetup{colorlinks=true,linkcolor=teal!70!black,citecolor=teal!70!black,urlcolor=teal!70!black,pdftitle={QAdapt: Noise-Adaptive Neural Pre-Decoding for Surface-Code Quantum Error Correction},pdfauthor={Ran Miao, Rui Luo, Xiaohan Shan, Xiaoming Sun}}
\title{\vspace{-1.5em}\textbf{QAdapt: Noise-Adaptive Neural Pre-Decoding\\for Surface-Code Quantum Error Correction}}
\author[1]{Ran Miao}
\author[1,2]{Rui Luo}
\author[1]{Xiaohan Shan}
\author[3]{Xiaoming Sun}
\affil[1]{Beijing Zhongke Liangshu Technology Co., Ltd, Beijing, China}
\affil[2]{Center for Quantum Information, Institute for Interdisciplinary Information Sciences, Tsinghua University, Beijing, China}
\affil[3]{Institute of Computing Technology, Chinese Academy of Sciences, Beijing, China}

\date{}
\begin{document}
\raggedbottom
\twocolumn[
\maketitle
\vspace{-1.6em}
\begin{abstract}
Fault-tolerant quantum computing relies on quantum error correction to protect logical information, requiring classical decoders that keep pace with syndrome generation. Heterogeneous, changing hardware noise and the distribution shift from simulation to experiment challenge the accuracy and robustness of fixed neural decoders. We present QAdapt, a noise-adaptive neural pre-decoding framework for surface-code quantum error correction. QAdapt learns local spatiotemporal correlations, continually adapts across noise environments, and passes the residual syndrome to a conventional global decoder. Online elastic weight consolidation limits forgetting as training moves between measurement, idle, CNOT, and Pauli-bias noise regimes. At distance nine, the HTNet pre-decoder reduces equivalent per-round logical error by 15.64\% relative to Ising-Fast, with 28.75\% fewer parameters and 9.07\% less matching-backend time. Online EWC reduces average forgetting by 76.52\% relative to sequential fine-tuning. With weights frozen, it reduces equivalent per-round logical error by 5.66\% relative to continued base-environment training on a fixed valid domain of 331 unseen compound-noise conditions; whole-sequence failure probability falls by 2.54\% over all 396 conditions. On native Willow records, the corresponding per-round reductions are 19.44\%, 17.47\%, and 17.95\% at 10, 30, and 50 rounds, without target-domain fine-tuning. These results show how adaptive neural pre-decoding can improve logical accuracy, preserve performance across noise regimes, and reduce residual matching workload within an established decoding pipeline.
\end{abstract}
\vspace{1.1em}
]

\section{Introduction}
Fault-tolerant quantum computing (FTQC) relies on quantum error correction (QEC) to protect logical information encoded in noisy physical qubits. QEC operates as a quantum--classical feedback loop: repeated stabilizer measurements generate syndrome data, classical decoders infer errors, and corrections or Pauli-frame updates track their effects~\citep{sivak2023realtime,sundaresan2023multiround}. Its scalability therefore depends on both the quality of quantum hardware and the accuracy, latency, and robustness of classical decoding. As code distances and circuit depths increase, the decoding stack must process growing syndrome streams without delaying logical operations~\citep{skoric2023parallel,caune2024realtime}.

Meeting this requirement on realistic hardware is challenging. Noise is heterogeneous and nonstationary: the relative contributions of measurement errors, entangling-gate faults, idle errors, and biased noise can vary across qubits and calibration cycles. Simulation-based training introduces a further mismatch because a simulated noise model cannot fully describe device-specific behavior. A decoder optimized for one distribution may consequently lose accuracy or require more computation under another, even when the quantum code is unchanged. Repeated retraining depends on representative data from each operating regime, and sequential updates can degrade performance on previously encountered regimes. Maintaining decoding performance under such changes is therefore a practical requirement for QEC~\citep{hockings2025noiseaware,bhardwaj2025drifting}.

Surface codes provide a leading setting for addressing these challenges because of their local stabilizer structure and favorable error thresholds~\citep{dennis2002topological,fowler2012surface}. Experiments have demonstrated logical-error suppression with increasing code distance and below-threshold quantum memories on Google's Willow processors~\citep{acharya2023suppressing,google2025willow}. These advances make the coupling between hardware noise and classical decoding increasingly consequential. Higher detector-event densities increase the workload presented to a decoder, while changes in error correlations can invalidate the assumptions or learned features on which its predictions depend.

Conventional global decoders, including minimum-weight perfect matching, enforce consistency across detector events and benefit from efficient implementations such as PyMatching~\citep{higgott2021pymatching,higgott2025sparse}. Their computational workload nevertheless grows with the active decoding problem, motivating a balance between logical accuracy and processing cost~\citep{demartiiolius2024review,delfosse2023tradeoff}. Neural decoders can learn correlated error structures that are difficult to encode explicitly~\citep{varbanov2023neural,bausch2024alphaqubit}, but their predictions depend on how well training covers the noise patterns encountered during operation.

Neural pre-decoding combines these capabilities in a modular pipeline. A learned model first predicts local corrections, and a conventional global decoder resolves the remaining detector events~\citep{chamberland2023localglobal,chamberland2026predecoders}. Removing locally recognizable errors can reduce residual syndrome density and backend workload while retaining global consistency. The effectiveness of this division of work depends on whether the neural component remains useful as noise changes. It must capture the relevant spatial and temporal correlations, adapt to new regimes, and retain what it has learned from earlier ones.

We present QAdapt, a noise-adaptive neural pre-decoding framework that addresses these requirements within a single inference network. Its HTNet pre-decoder combines spatial, temporal, and joint features using input-dependent channel weights. Online elastic weight consolidation (EWC) constrains parameter changes according to their importance for earlier noise environments~\citep{kirkpatrick2017ewc,schwarz2018progress}. In the controlled adaptation sequence studied here, training moves between measurement, idle, CNOT, and Pauli-bias noise regimes. The resulting weights are frozen for evaluation on unseen compound noise and native Willow records. The residual syndrome continues to be processed by the same global matching backend.

This separation allows the learned component to adapt across noise regimes without changing the quantum code or the deployed inference graph. It also connects two practical objectives: preserving logical accuracy as noise changes and reducing the residual workload passed to an established decoder. We evaluate these objectives through paired architecture comparisons, continual-learning controls, and frozen transfer tests.

The main contributions of this work are as follows:
\begin{enumerate}
  \item \textbf{Noise-adaptive pre-decoding.} QAdapt supports sequential adaptation across physically interpretable noise environments. With matched training-update budgets, Online EWC reduces average forgetting by 76.52\% relative to sequential fine-tuning, preserving more of the performance acquired on earlier regimes.
  \item \textbf{Decoder-compatible integration.} HTNet combines conditional spatiotemporal features with residual global matching. A paired distance-nine evaluation with over two million shots shows a 15.64\% reduction in equivalent per-round logical error relative to Ising-Fast, using 28.75\% fewer parameters. Factorial ablations identify the contributions of post-fusion mixing and raw-input access.
  \item \textbf{Robustness under distribution shift.} Frozen evaluations cover 396 unseen compound-noise conditions and native Willow records at 10, 30, and 50 rounds. Relative to continued base-environment training, Online EWC reduces equivalent per-round logical error by 5.66\% on the fixed 331-condition valid domain and by 17.47--19.44\% on Willow, without target-domain fine-tuning. Whole-sequence failure probability decreases by 2.54\% over the complete compound-noise grid.
  \item \textbf{Reduced backend workload.} In the same distance-nine architecture comparison, HTNet reduces residual detector density by 14.14\% and matching-backend time by 9.07\% relative to Ising-Fast. These measurements link improved local error correction to a smaller residual decoding problem.
\end{enumerate}

\section{Related work}
Minimum-weight perfect matching is a standard surface-code decoding method, and sparse blossom provides a fast implementation in PyMatching~\citep{higgott2021pymatching,higgott2025sparse}. Local--global decoding combines a preliminary correction stage with a global decoder~\citep{chamberland2023localglobal}. Neural versions learn this preliminary stage from simulated faults, including recent fast AI-based pre-decoders~\citep{chamberland2026predecoders}. QAdapt uses the same modular principle and compares against an Ising-Fast implementation trained in the same experimental pipeline.

Neural decoders can learn correlations that are difficult to encode in a fixed matching graph. AlphaQubit demonstrates high-accuracy learned decoding on quantum-processor data~\citep{bausch2024alphaqubit}. QuantiSpect introduces a lightweight three-dimensional pre-decoder with spatial, temporal, and joint convolutional branches~\citep{gao2026quantispect}. HTNet builds on that factorization. Its conditional branch fusion, grouped post-fusion mixing, and raw-input skip are used here with a selected channel-gating configuration. Our factorial study isolates the mixer, axis gate, and input skip within HTNet, while Ising-Fast provides the architecture-level comparison.

Noise-aware decoding and estimation of drifting noise address the mismatch between a decoder's assumptions and a device's operating conditions~\citep{hockings2025noiseaware,bhardwaj2025drifting}. QAdapt instead studies how a neural pre-decoder retains useful parameters across a controlled sequence of simulated noise regimes. EWC penalizes changes to parameters important for past tasks~\citep{kirkpatrick2017ewc}; its online form consolidates this information into a fixed-size state~\citep{schwarz2018progress}. We use that established regularizer and evaluate its consequences for surface-code pre-decoding, including comparisons with replay and mixed-task training. The continual-learning metrics follow the distinction between task acquisition and backward transfer~\citep{lopezpaz2017gem}.

\section{QAdapt framework}
\label{sec:method}
\subsection{Residual decoding}
Let $\bm s$ denote a detector tensor and $f_\theta(\bm s)$ the neural correction logits. After thresholding, the predicted binary correction $\hat{\bm e}$ changes the detector pattern through the detector--correction incidence map $H$:
\begin{equation}
\bm s_{\mathrm{res}}=\bm s\oplus H\hat{\bm e}.
\label{eq:residual}
\end{equation}
If $L$ maps a correction to its logical parity and $\mathcal D$ is the residual matching decoder, the final logical prediction is
\begin{equation}
\hat{\ell}=L\hat{\bm e}\oplus\mathcal D(\bm s_{\mathrm{res}}).
\label{eq:logical}
\end{equation}
Thus, both the neural correction and the residual matching result contribute to the final decision. Comparisons use the same detector graph and prior within an evaluation condition. The neural network changes the observed detector events, while PyMatching completes the residual problem.

\begin{figure*}[!t]
\centering
\includegraphics[width=\textwidth]{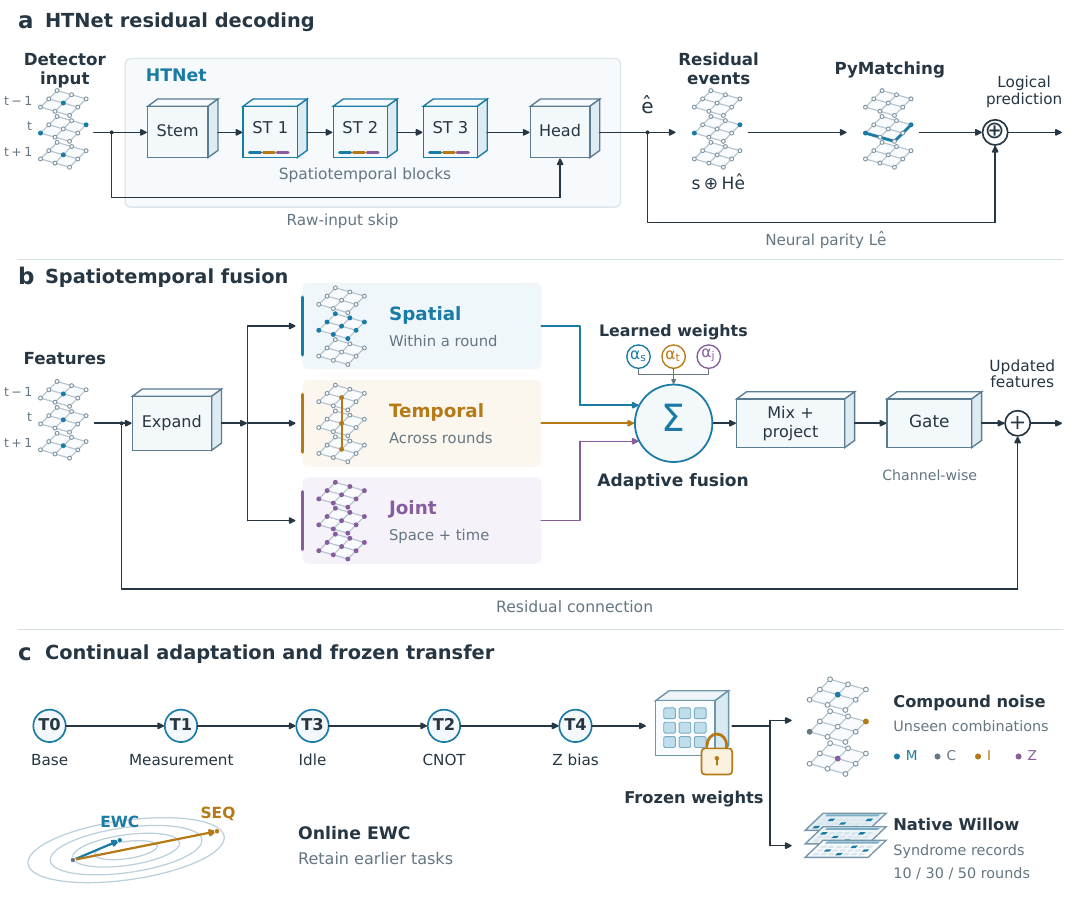}
\caption{\textbf{QAdapt combines neural pre-decoding with continual adaptation.} (a) HTNet comprises a stem, three spatiotemporal (ST) blocks, and an output head with a raw-input skip. The predicted correction and residual matching contribute their logical parities to the final prediction. (b) Each ST block combines spatial, temporal, and joint features using input-dependent, per-channel weights. Mixing, projection, gating, and dropout precede the residual addition. The module icons are schematic; layer specifications are given in the Methods and Appendix~\ref{app:training}. (c) Online EWC regularizes sequential adaptation; the resulting weights are frozen for evaluation on unseen compound noise and native Willow records. The inset illustrates parameter drift relative to SEQ. M, C, I, and Z denote measurement, CNOT, idle, and Z-bias noise.}
\label{fig:method}
\end{figure*}

\subsection{Conditional spatiotemporal features}
The input has four channels on a space--time grid. A $3\times3\times3$ stem maps it to 112 channels. Three residual fusion blocks then process the features. Each block applies group normalization and a pointwise projection to 168 channels, followed by three paths:
\begin{align}
\bm z_s&=\operatorname{DWConv}_{1\times3\times3}(\bm u),\nonumber\\
\bm z_t&=\operatorname{DWConv}_{3\times1\times1}(\bm u),\nonumber\\
\bm z_j&=\operatorname{GConv}_{3\times3\times3,g=6}
          (\operatorname{GN}(\bm u)).
\label{eq:branches}
\end{align}
The first coordinate is time. The spatial and temporal convolutions are depthwise; the joint path uses six channel groups. This allocation retains a joint space--time path without requiring dense three-dimensional convolution in every branch.

Global average pooling produces a descriptor for each branch. A small network maps the concatenated descriptors to three logits per channel, and a softmax normalizes them across branches. For channel $c$,
\begin{equation}
\bm z_c=3\sum_{k\in\{s,t,j\}}\alpha_{k,c}(\bm s)\bm z_{k,c},
\qquad \sum_k\alpha_{k,c}=1.
\label{eq:fusion}
\end{equation}
The final weight-producing layer is initialized to zero, so the block starts from equal branch weights; the factor of three then recovers the scale of a branch sum. A grouped $1\times1\times1$ mixer, projection to 112 channels, channel gate, and dropout precede the identity residual addition. The output head normalizes the final features, concatenates the original four-channel input, and applies pointwise maps $116\rightarrow112\rightarrow4$.

The selected configuration, A101, uses post-fusion mixing, ordinary channel gating, and the raw-input skip. It contains 650,332 parameters. In the three-bit configuration name, the bits indicate mixer, axis gate, and input skip, respectively. Adaptive branch fusion is present in every tested configuration; the axis-gate ablation replaces the ordinary channel gate. The architecture accepts different spatial and temporal dimensions with the same weights. Global pooling and normalization use statistics from the full input tensor.

\subsection{Continual adaptation with a fixed-size state}
Starting from base weights $\theta_0^\star$, the model learns one noise environment at a time. For stage $t$, Online EWC minimizes
\begin{equation}
\mathcal L_t(\theta)=\mathcal L_{\mathrm{BCE},t}(\theta)
+\frac{\lambda}{2}\sum_i\bar F_{t-1,i}
(\theta_i-\theta_{t-1,i}^\star)^2,
\label{eq:ewc}
\end{equation}
where the binary cross-entropy uses simulated correction targets. After the stage, the empirical diagonal importance estimate is
\begin{equation}
F_{t,i}=\frac{1}{N_F}\sum_{n=1}^{N_F}
\left[\partial_{\theta_i}\mathcal L_{\mathrm{BCE}}
(\theta_t^\star;x_n,y_n)\right]^2,
\end{equation}
and the stored state is updated as
\begin{equation}
\bar F_t=\gamma\bar F_{t-1}+F_t,
\qquad \theta^\star\leftarrow\theta_t^\star.
\end{equation}
We use $\lambda=316.227766$, $\gamma=1$, and $N_F=4,096$. The estimate averages squared per-sample gradients, rather than squaring a batch-averaged gradient. The initial importance estimate is computed at the T0 endpoint. Subsequent training retains one parameter anchor and one accumulated diagonal estimate, requiring $O(P)$ state for $P$ parameters and no stored historical examples. The regularizer changes training only; the inference graph and parameter count remain unchanged.

\section{Evaluation design}
\label{sec:evaluation}
\subsection{Noise environments and training controls}
All models train on distance $d=9$ with $r=9$ rounds. The base environment T0 is a 25-parameter circuit-level Pauli model that separates preparation, measurement, idle, and two-qubit faults. It has preparation probabilities $0.001$, measurement probabilities $0.01$, and a total CNOT Pauli-fault probability of approximately $0.01$. Individual coefficients and idle conventions appear in Appendix~\ref{app:training}.

The continuation sequence changes one noise family per task: T1 increases measurement noise by $4\times$, T3 increases idle noise by $4\times$, T2 increases CNOT noise by $1.5\times$, and T4 increases the Z-bias parameters by $1.5\times$, all relative to T0. These environments and their order were selected as a controlled adaptation stress test using in-distribution diagnostics. They represent distinct operating regimes, not a time series measured from a drifting device.

Architecture comparisons use the validation-best T0 checkpoints: epoch 97 for HTNet and epoch 87 for Ising-Fast. Continual-learning comparisons instead start from the fixed epoch-100 checkpoint of each architecture. We denote this starting point C0 and the four continuation endpoints C1--C4. Checkpoint labels indicate training progress; task labels identify evaluation noise distributions.

Each sequential method receives 15,360 optimizer updates per task, or 61,440 continuation updates in total, with an effective batch size of 4,096. SEQ uses unregularized sequential fine-tuning. Online EWC adds Eq.~\eqref{eq:ewc}. Replay replaces 25\% of each batch with stored historical examples under a persistent-state memory budget matched to EWC. T0-only receives the same continuation budget on T0. Mix-Total samples T0--T4 uniformly within the same total update budget. Mix-Exposure adds 25\% more updates to match per-task exposure across the five environments. The latter two controls have simultaneous access to all five training distributions.

Hyperparameters were selected on HTNet using separate in-distribution validation and transferred to Ising-Fast. The compound-noise and Willow evaluations do not select hyperparameters or checkpoints. Each training configuration has one realized training seed; the statistical intervals below quantify evaluation-shot uncertainty conditional on those trained weights.

\subsection{Paired evaluation domains}
\textbf{Architecture.} At nine rounds, the base-checkpoint comparison covers $d\in\{5,7,9\}$ with 65,536 shots per logical basis and distance. A higher-statistics $d=9$ evaluation uses 1,048,576 shots per basis. The full factorial ablation uses eight configurations at $d=r=9$, also with 1,048,576 shots per basis. A length extension covers $r\in\{3,5,9,17,33\}$ at each of the three architecture-evaluation distances, with 10,000 shots per basis and condition.

\textbf{Continual learning.} Each checkpoint is evaluated on all five tasks T0--T4 with 65,536 shots per basis and task. These measurements separate immediate acquisition, historical retention, and terminal average performance.

\textbf{Unseen compound noise.} Frozen terminal weights are evaluated on all pairwise, three-way, and four-way combinations of measurement (M), CNOT (C), idle (I), and Z-bias (Z) changes. The 11 combinations, nine multipliers $\{1.2,1.5,2,2.5,3,3.5,4,4.5,5\}$, and four distances $\{3,5,7,9\}$ form 396 nine-round conditions, each with 10,000 shots per basis. A physical parameter shared by several selected axes is scaled once. Figure~\ref{fig:inputdensity} shows the input syndrome densities over this complete grid. A separate length-transfer grid uses M+I, C+Z, and M+C+I+Z at $1.5\times$, the same four distances, and five sequence lengths from 3 to 33 rounds: 60 conditions in total.

\textbf{Willow.} We evaluate native 10-, 30-, and 50-round records from the released Willow surface-code data~\citep{google2025willow}. Each length contains the same set of 28 patch/basis cases: 18 at $d=3$, eight at $d=5$, and two at $d=7$. Each case contains 50,000 shots, giving 4.2 million records across the three lengths. Each length uses its own native records and logical labels. No hardware-data fine-tuning is performed. All methods share the records within each case, and the matching baseline uses an uncorrelated SI1000 prior.

\subsection{Metrics and uncertainty}
\label{sec:metrics}
The whole-sequence logical failure probability is $p=N_{\rm err}/N$. We also report the equivalent per-round logical error rate (LER)
\begin{equation}
q=\frac{1-(1-2p)^{1/r}}{2},\qquad 0\leq p<\tfrac12.
\label{eq:q}
\end{equation}
This is the inverse of the independent logical-flip relation $p=[1-(1-2q)^r]/2$. For correlated noise and hardware records, $q$ is an equivalent rate derived from the full-sequence measurement.

We transform each condition/basis separately, average within each distance, and then weight distances equally:
\begin{equation}
\bar q=\frac{1}{|\mathcal D|}\sum_{d\in\mathcal D}
\frac{1}{|S_d|}\sum_{c\in S_d}
\frac{1}{B_c}\sum_{b=1}^{B_c}q_{dcb}.
\label{eq:average}
\end{equation}
Simulation conditions have two logical bases; each Willow patch/basis case has one. Relative improvement is the ratio of the resulting averages,
\begin{equation}
G_q=100\left(1-\frac{\bar q_{\rm candidate}}{\bar q_{\rm baseline}}\right)\%,
\label{eq:gain}
\end{equation}
not the average of per-condition relative changes. The same weighting defines $\bar p$ and $G_p$.

For the compound-noise grid, the original evaluation fixed a common domain on which all evaluated neural models in both bases have $p<0.5$. We preserve that mask for every reported $q$ comparison: 331 conditions, comprising 99, 88, 77, and 67 at distances 3, 5, 7, and 9. The full-grid $p$ analysis retains all 396 conditions. This distinction keeps the nonlinear rate conversion from changing the comparison domain as the noise becomes stronger.

Let $Q_{s,j}$ be the mean $q$ on task $j$ at checkpoint C$s$, and let $\tau_j$ be the checkpoint at which task $j$ was acquired. With $\mathcal H=\{j:\tau_j<4\}$, average forgetting and backward transfer are
\begin{align}
\mathrm{AF}&=\frac{1}{|\mathcal H|}\sum_{j\in\mathcal H}
\left[Q_{4,j}-\min_{\tau_j\leq s<4}Q_{s,j}\right],\label{eq:af}\\
\mathrm{BWT}&=\frac{1}{|\mathcal H|}\sum_{j\in\mathcal H}
\left[Q_{\tau_j,j}-Q_{4,j}\right].\label{eq:bwt}
\end{align}
The historical tasks are T0, T1, T3, and T2; the final task T4 is excluded. Negative forgetting is retained when a task improves beyond its earlier best. We report AF and BWT in percentage points of $q$, distinct from relative percentage reductions. Immediate acquisition compares a task immediately before and after its own training stage.

Paired methods see identical shots. Approximate 95\% intervals use paired error counts and the delta method through Eqs.~\eqref{eq:q}--\eqref{eq:gain}. The covariance term accounts for errors made on the same shots by both methods. Backend time measures residual \texttt{matcher.decode} calls only: after 50 warm-up calls, five serial repetitions are reduced to the median per basis, averaged over bases, and divided by the number of rounds.

\section{Results}
\subsection{Improved accuracy and residual workload}
\label{sec:archresults}
\begin{figure*}[!t]
\centering
\includegraphics[width=\textwidth]{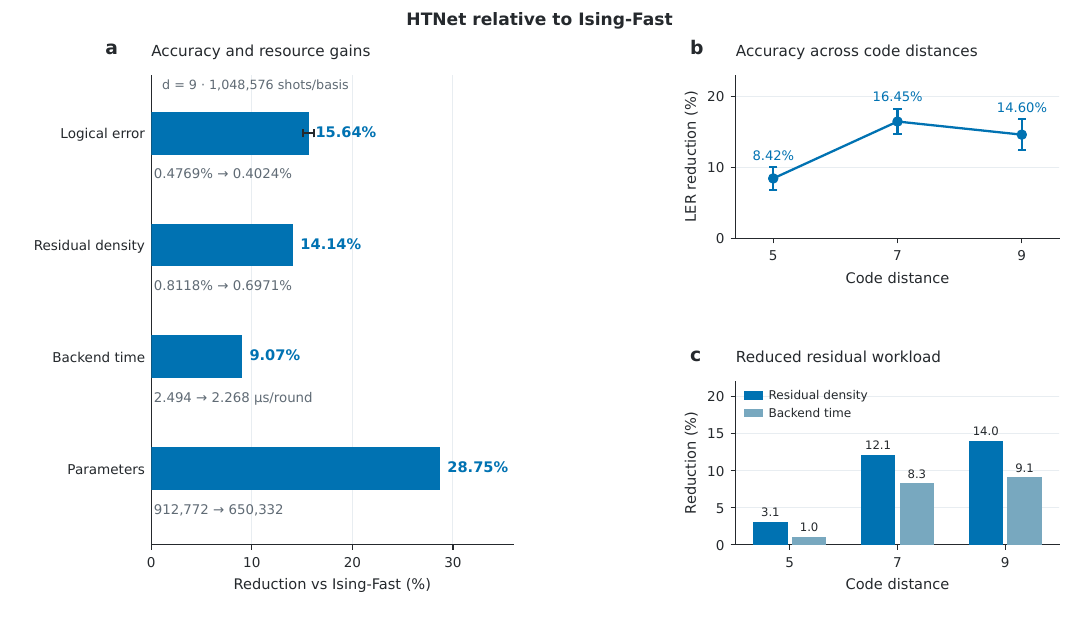}
\caption{\textbf{HTNet improves accuracy and reduces resource requirements relative to Ising-Fast.} (a) Relative reductions in per-round LER, residual detector density, matching-backend time, and parameter count. Absolute values below each bar read Ising-Fast $\rightarrow$ HTNet. Accuracy and workload measurements use the independent $d=r=9$ evaluation with 1,048,576 shots per basis; the LER whisker is a paired 95\% interval, while resource bars are point estimates. (b) LER reductions with paired 95\% intervals at $d=5,7,9$, using 65,536 shots per basis. (c) Residual-density and matching-backend reductions in the same distance sweep. All timing concerns residual matching. Complete factorial evidence appears in Fig.~\ref{fig:ablationdetail}.}
\label{fig:architecture}
\end{figure*}

At nine rounds, HTNet reduces per-round LER relative to Ising-Fast by 8.42\%, 16.45\%, and 14.60\% at distances 5, 7, and 9, respectively (Fig.~\ref{fig:architecture}b). The higher-statistics distance-nine evaluation resolves a 15.64\% reduction, with a paired 95\% interval of $[15.10,16.18]\%$: mean $q$ decreases from 0.4769\% to 0.4024\%. These weights were trained at a single distance and sequence length.

The residual event density falls from 0.8118\% to 0.6971\%, a 14.14\% reduction, and residual matching time falls from 2.494 to 2.268~$\mu$s per round, a 9.07\% reduction (Fig.~\ref{fig:architecture}a). HTNet uses 28.75\% fewer parameters than the 912,772-parameter Ising-Fast implementation.

The factorial comparison identifies where changes within HTNet help (Fig.~\ref{fig:ablationdetail}). Enabling the post-fusion mixer reduces nine-round failure probability by 0.1457 percentage points on average across the other factors. The raw-input skip reduces it by 0.0120 percentage points; its Holm-adjusted $p$ value is 0.0386. The mean axis-gate effect is indistinguishable from zero. Moreover, the axis-gate/input-skip interaction increases failure probability by 0.0590 percentage points, showing that individually plausible additions need not combine favorably. A101 has the lowest observed mean among the eight variants and is used in subsequent experiments. 

The length extension supports accuracy and backend gains beyond the training length. At 33 rounds, LER reductions at distances 5, 7, and 9 are 10.58\%, 14.32\%, and 10.99\%; backend-time reductions are 6.91\%, 13.94\%, and 12.49\%. At three rounds the architecture comparison can reverse, as shown by the complete length grid in Appendix~\ref{app:lengths}. Thus the observed advantage depends on the temporal context available to the pre-decoder.

\subsection{Online EWC reduces historical forgetting}
\label{sec:clresults}
\begin{figure*}[!t]
\centering
\includegraphics[width=\textwidth]{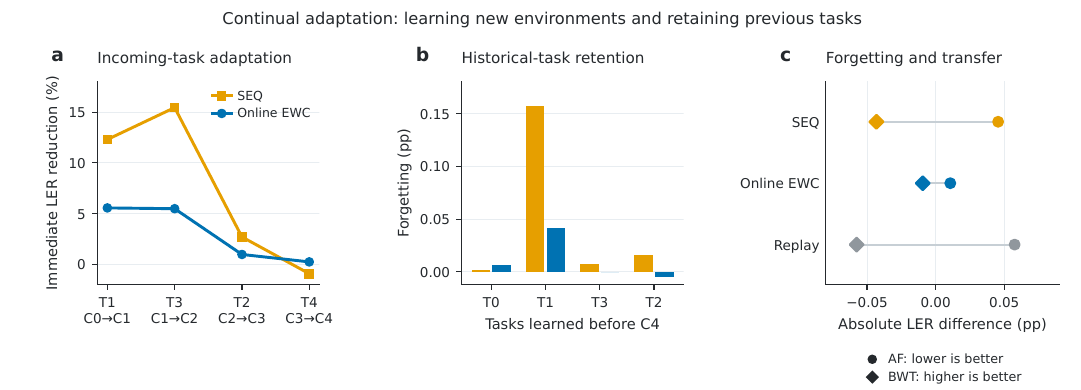}
\caption{\textbf{New-task acquisition and historical retention.} (a) Relative LER reduction on each incoming task, comparing the checkpoints immediately before and after its training stage. (b) Final error above the historical best for each previously learned task; negative values indicate further improvement. (c) Average forgetting (AF, circles) and backward transfer (BWT, diamonds) for SEQ, Online EWC, and Replay, in absolute percentage points. Lower AF and higher BWT are better. Each task/checkpoint evaluation uses 65,536 shots per basis. The trajectories summarize one training realization per setting; the 76.52\% forgetting reduction is a point estimate over the four historical tasks.}
\label{fig:continual}
\end{figure*}

SEQ rapidly adapts to the first three new environments: immediate LER reductions on T1, T3, and T2 are 12.31\%, 15.46\%, and 2.70\%. Online EWC produces smaller changes of 5.56\%, 5.48\%, and 0.97\% (Fig.~\ref{fig:continual}a). On T4, the changes are $-0.93\%$ for SEQ and $0.23\%$ for EWC. The regularizer therefore moderates the movement toward each current environment.

Its main in-distribution benefit appears after further training. Average forgetting decreases from 0.0455 to 0.0107 percentage points, a 76.52\% reduction; BWT improves from $-0.0432$ to $-0.0094$ percentage points. The measurement task accounts for the largest individual difference. After SEQ learns T1, training on the later environments raises T1's final error by 0.1570 percentage points above its historical best; EWC limits this increase to 0.0420 percentage points (Fig.~\ref{fig:continual}b,c).

Retention does not coincide with the lowest terminal task average. SEQ ends at 1.0715\% mean in-distribution LER and EWC at 1.0871\%, a 1.45\% relative difference in favor of SEQ. Replay ends at 1.0966\% with greater forgetting, while Mix-Total and Mix-Exposure end at 1.1757\% and 1.1424\%. Table~\ref{tab:terminal} places these results next to frozen transfer performance. The EWC benefit is preservation across the trajectory together with stronger transfer, rather than a uniform reduction of every in-distribution metric.

\begin{table*}[t]
\centering\small
\caption{\textbf{Terminal performance across training and transfer domains.} All rows use HTNet. In-distribution (ID) LER averages T0--T4; AF and BWT are absolute percentage-point differences. OOD LER uses the fixed 331-condition domain; Willow uses the 28 ten-round patch/basis cases with equal distance weights. Lower LER and AF are better; higher BWT is better. Mix-Exposure receives 25\% more continuation updates. The transfer columns are equivalent per-round rates, not whole-sequence failure probabilities.}
\label{tab:terminal}
\input{data/terminal_table.tex}
\end{table*}

\subsection{Frozen transfer to unseen compound noise}
\label{sec:oodresults}
\begin{figure*}[!t]
\centering
\includegraphics[width=\textwidth]{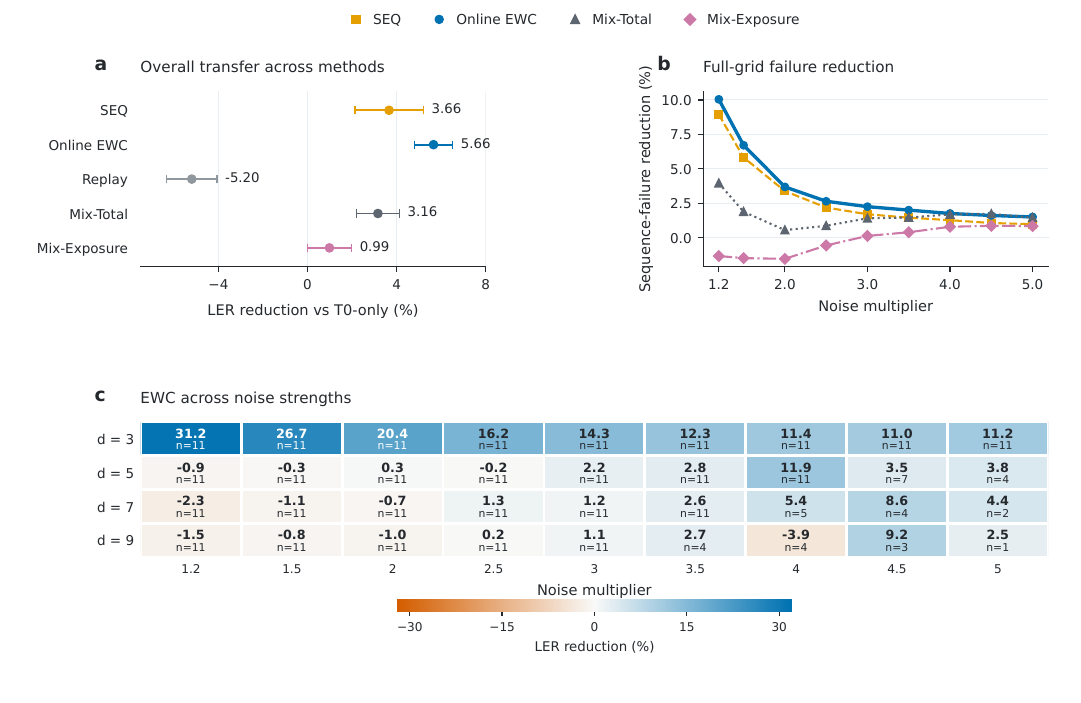}
\caption{\textbf{Frozen transfer across adaptation controls and noise strengths.} (a) Per-round LER reduction relative to HTNet T0-only for five adaptation controls, with paired 95\% intervals on the fixed 331-condition validity domain. (b) Whole-sequence failure reduction across all 396 conditions, including 44 per multiplier. (c) EWC LER gains at each distance and multiplier; blue indicates improvement and orange deterioration. Each cell gives the relative reduction and valid-condition count $n$, out of 11 noise combinations. Each condition uses 10,000 shots per basis. Figure~\ref{fig:ooddetail} gives noise-family results and the SEQ strength map; Fig.~\ref{fig:transfercontrols}a gives the distance-resolved control comparison.}
\label{fig:ood}
\end{figure*}

On the fixed common domain, EWC reduces mean per-round LER from 6.0848\% for T0-only to 5.7404\%, a relative reduction of 5.66\% with a paired 95\% interval of $[4.81,6.51]\%$ (Fig.~\ref{fig:ood}a). SEQ reaches 5.8620\%, a 3.66\% reduction. EWC is 2.07\% better than SEQ on the same domain, with interval $[0.62,3.52]\%$. The EWC gains relative to Mix-Total and Mix-Exposure are 2.58\% and 4.72\%, respectively. Exposure to several environments alone therefore does not reproduce the transfer result of the selected sequential regularization protocol. With EWC held fixed across architectures, HTNet also has a 1.93\% lower mean OOD LER than Ising-Fast on the same domain.

The improvement is not confined to one compound-noise family. All 11 family-level mean gains are positive; the largest is 8.87\% for combined measurement and idle changes (Fig.~\ref{fig:ooddetail}a). EWC improves 230 of the 331 individual conditions. Distance-resolved mean gains decrease from 13.98\% at $d=3$ to 4.59\%, 2.51\%, and 0.84\% at $d=5,7,9$ (Fig.~\ref{fig:transfercontrols}a). The distance-nine interval spans zero, so the overall gain does not establish a uniform improvement across code sizes. Figure~\ref{fig:ood}c resolves this dependence further across noise strengths, retaining the valid-condition count in each cell.

On the complete grid, EWC reduces mean whole-sequence failure probability from 0.31967 to 0.31155, a 2.54\% reduction with interval $[2.44,2.64]\%$. It improves 275 of all 396 conditions. The full-grid reduction remains positive at every tested multiplier, although the gain diminishes as the noise strengthens (Fig.~\ref{fig:ood}b). This analysis includes the high-noise conditions where the equivalent per-round rate is undefined.

Length transfer provides a distinct test. For the three fixed $1.5\times$ compound-noise families, EWC lowers the distance-balanced per-round rate relative to T0-only at every tested length. The gains are 8.26\%, 6.69\%, and 4.83\% at 9, 17, and 33 rounds. At the two longer lengths, however, the overall gain is driven by $d=3$ while the $d=5,7,9$ means slightly worsen. Appendix~\ref{app:oodlength} displays this heterogeneity. The length experiment thus supports aggregate transfer under its specified weighting, with a different distance profile from native hardware transfer.

\begin{figure*}[!t]
\centering
\includegraphics[width=\textwidth]{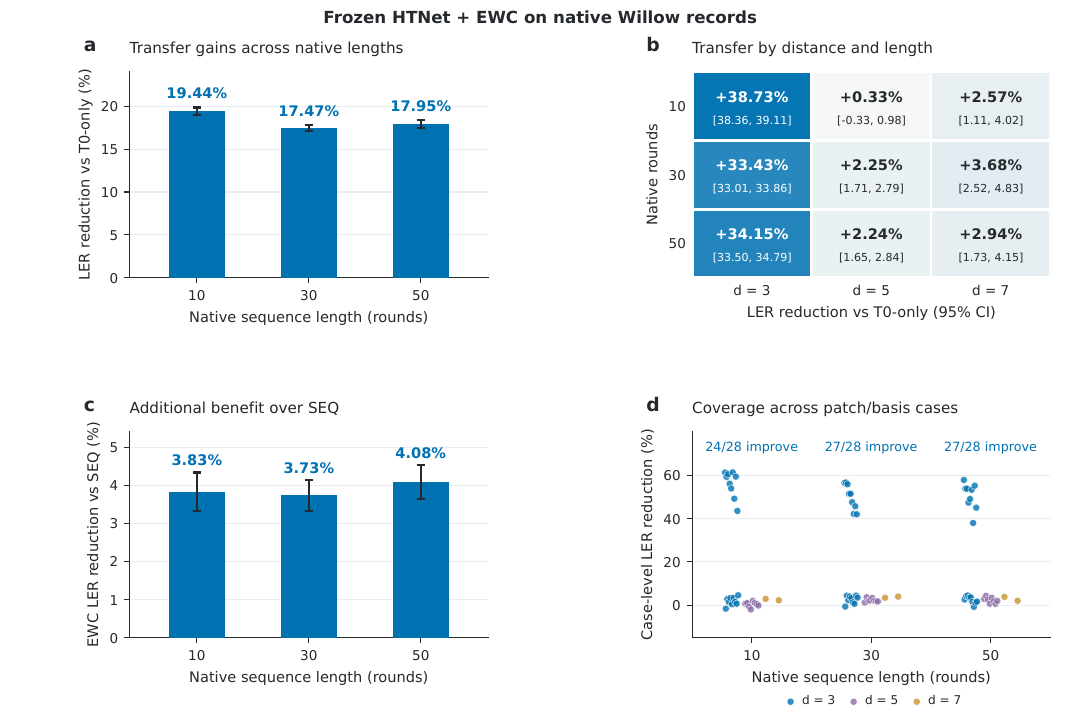}
\caption{\textbf{EWC improves frozen transfer across native Willow sequence lengths.} (a) Distance-balanced per-round LER reduction relative to HTNet T0-only at 10, 30, and 50 rounds. (b) The same comparison by distance and native length, with paired 95\% intervals in brackets. (c) EWC reduction relative to SEQ on the same records. Whiskers in (a,c) are paired 95\% intervals. (d) Every patch/basis case relative to T0-only; color identifies distance, and labels count cases with a lower point-estimate LER. All 28 cases per length are included, including reversals; horizontal offsets separate points and have no numerical meaning. Each case has 50,000 shots. Models were trained at nine rounds and remain frozen. Absolute neural-method rates and mixed-training comparisons appear in Fig.~\ref{fig:transfercontrols}.}
\label{fig:willow}
\end{figure*}

\subsection{Frozen transfer to native Willow records}
\label{sec:willowresults}
The weights trained on nine-round simulated circuits transfer directly to Willow records (Fig.~\ref{fig:willow}). At ten rounds, EWC lowers the distance-balanced per-round LER from 1.2541\% to 1.0104\%, a 19.44\% reduction with interval $[19.01,19.86]\%$. It improves 24 of 28 patch/basis cases. SEQ yields a 16.22\% reduction, and EWC improves on SEQ by 3.83\% and on Mix-Total by 1.40\%.

The effect persists when the native sequence length increases. At 30 and 50 rounds, EWC reduces LER relative to T0-only by 17.47\% and 17.95\%, respectively. Its advantages over SEQ are 3.73\% and 4.08\%, with paired intervals above zero. EWC and Mix-Total are effectively tied at these lengths: their relative differences are $-0.05\%$ and $-0.15\%$, with intervals spanning zero (Fig.~\ref{fig:transfercontrols}b).

The distance decomposition sharpens the result. At ten rounds, EWC gains are 38.73\%, 0.33\%, and 2.57\% for $d=3,5,7$. At fifty rounds they are 34.15\%, 2.24\%, and 2.94\%, with positive paired intervals at all three distances. Twenty-seven of the 28 fifty-round cases improve, including all eight $d=5$ cases. Equal distance weighting prevents the 18 distance-three cases from dominating through their greater number, although the larger distance-three effect still contributes most to the mean reduction.

\section{Discussion}
Conditional feature extraction and parameter consolidation act at different points in the decoding pipeline. The former changes how the network represents detector structure; the latter constrains how its weights change between noise regimes. The measurements support their use together: the architecture improves base-task accuracy, and the selected EWC trajectory combines better historical retention with lower transfer error.

The difference between terminal ID performance and transfer is informative. SEQ has a slightly lower final ID mean than EWC, yet EWC performs better on the main OOD grid and on Willow. Selecting a continual pre-decoder solely by its final average on the training environments would miss this distinction. A useful selection protocol should therefore report the acquired-task trajectory and hold out noise combinations and hardware records from model selection.

HTNet reduces both parameter count and the residual workload presented to matching. At distance nine, the 14.14\% reduction in residual event density accompanies a 9.07\% reduction in matching-backend time. This connection makes residual sparsity a useful companion metric to logical accuracy when selecting a pre-decoder. The use of global pooling and normalization also makes temporal context an explicit architectural choice, consistent with the observed sequence-length dependence.

Replicated training would measure sensitivity to initialization, data generation, and optimization. A sequence drawn from measured device drift would test whether the controlled simulated transitions capture deployment behavior. Both extensions address questions beyond evaluation-shot precision and would help determine how to select adaptation schedules for a changing processor.

\section{Conclusion}
QAdapt combines conditional spatiotemporal pre-decoding with fixed-state continual adaptation. In the reported experiments, HTNet lowers distance-nine logical error by 15.64\% with 28.75\% fewer parameters, and Online EWC lowers average forgetting by 76.52\% relative to sequential fine-tuning. Frozen weights improve on continued single-environment training across the compound-noise grid and native Willow records through fifty rounds. The resulting evidence supports training pre-decoders for retained competence across noise environments, alongside optimizing their base-distribution features. Further work should test these adaptation gains across independent training runs and measured device-drift sequences.

\section*{Acknowledgments}
We sincerely thank Fusheng Chen (Hefei National Laboratory, University of Science and Technology of China) and Riling Li (Institute of Computing Technology, Chinese Academy of Sciences) for their thoughtful and valuable discussions of this work.

\section*{Data and code availability}
The numerical data underlying the figures, paired-error summaries, and manuscript sources accompany this manuscript. Pretrained QAdapt v2 weights, comparison checkpoints, inference scripts, and the required code patch are available on Hugging Face at \url{https://huggingface.co/Qhub-AI/QAdapt}, under the \href{https://huggingface.co/Qhub-AI/QAdapt/tree/v2}{\texttt{v2}} release. The versioned model files are stored in the \texttt{v2/} directory. The Willow source records are available with the original experiment at \url{https://doi.org/10.5281/zenodo.13273331}. The accompanying analysis archive records the provenance of the simulation evaluations. The complete simulator and training codebase is not included in that archive.

\begingroup
\small
\setlength{\bibsep}{2.5pt}
\bibliographystyle{unsrtnat}
\bibliography{references}
\endgroup
\clearpage
\onecolumn
\appendix
\setcounter{figure}{0}
\renewcommand{\thefigure}{S\arabic{figure}}
\setcounter{table}{0}
\renewcommand{\thetable}{S\arabic{table}}

\section{Training and implementation details}
\label{app:training}
\subsection{Circuit noise and task construction}
T0 specifies preparation-X and preparation-Z probabilities of $0.001$ each and measurement-X and measurement-Z probabilities of $0.01$ each. During CNOT-layer idles, the X, Y, and Z channels each have probability $0.000333$. The idle-SPAM parameters are $0.000667$ per Pauli channel. Each of the 15 non-identity two-qubit Pauli channels following a CNOT has probability $0.000667$, giving total probability $0.010005$. These coefficients are used directly, without training-noise upscaling. The simulator applies its preparation/reset idle convention; data idles during ancilla measurement are omitted. T1--T4 scale the corresponding parameter subsets relative to this fixed base, rather than cumulatively scaling the preceding task's noise.

The training labels use simulated faults and the pipeline's homological-equivalence reduction, as in the underlying neural pre-decoding workflow. Both logical bases contribute training samples. Binary cross-entropy supervises the four correction channels. The inference decision uses a probability threshold of $0.5$ (zero logit), followed by the detector and logical-parity updates in Eqs.~\eqref{eq:residual} and \eqref{eq:logical}. Detector simulation uses Stim-compatible circuit-level fault models~\citep{gidney2021stim}.

Task selection used 16 single-axis candidate environments and in-distribution feature and gradient diagnostics to construct a controlled stress sequence. The selected order alternates environments with competing update directions. The Z-bias subset comprises X-basis preparation and measurement flips, the Z component of each idle channel, and CNOT Pauli terms IZ, XZ, YZ, ZI, ZX, ZY, and ZZ. Shared coefficients are scaled once in compound environments.

The final regularization setting was chosen under a 2\% relative terminal-ID tolerance against SEQ and a 5\% current-task tolerance. Fisher estimation used 4,096 samples following calibration against a 65,536-sample reference, with cosine similarity and norm agreement as convergence checks. These choices were frozen before the transfer evaluations reported here.

\subsection{Architecture implementation}
The three pooled branch descriptors have 504 channels. Their weighting network uses pointwise maps $504\rightarrow28\rightarrow504$ with GELU, followed by a softmax over the three branches for each of the 168 channels. The post-fusion pointwise mixer uses six groups. The channel gate applies global average pooling and pointwise maps $112\rightarrow28\rightarrow112$, with GELU between the maps and a final sigmoid. The resulting weights are broadcast over the time and spatial axes.

\subsection{Optimization and persistent state}
Training uses the Lion optimizer. Online EWC maintains a parameter anchor and an accumulated diagonal Fisher estimate, while Replay stores historical examples under the same persistent-state memory budget. Both methods retain the same inference architecture.

\begin{table}[H]
\centering\small
\caption{\textbf{Complete HTNet factorial ablation.} $M$, $A$, and $S$ indicate mixer, axis gate, and input skip. Rates average X and Z after their respective conversion. Each basis has 1,048,576 shots at $d=r=9$.}
\label{tab:ablation}
\input{data/ablation_table.tex}
\end{table}

\section{Paired statistics and reproducibility}
\label{app:statistics}
For candidate $a$ and baseline $b$ on the same $N$ shots, let $n_{11}$ count joint errors, $n_{10}$ candidate-only errors, and $n_{01}$ baseline-only errors. Then
\begin{equation}
\hat p_a=(n_{11}+n_{10})/N,\qquad
\hat p_b=(n_{11}+n_{01})/N.
\end{equation}
The estimated variance and covariance of the sample proportions are
\begin{align}
\widehat{\operatorname{Var}}(\hat p_a)&=\frac{\hat p_a(1-\hat p_a)}{N-1},\\
\widehat{\operatorname{Cov}}(\hat p_a,\hat p_b)&=
\frac{n_{11}/N-\hat p_a\hat p_b}{N-1}.
\end{align}
For $g_r(p)=[1-(1-2p)^{1/r}]/2$, the derivative is
\begin{equation}
g_r'(p)=\frac{1}{r}(1-2p)^{1/r-1}.
\end{equation}
Multiplying the probability variances by the squared derivatives, and the covariance by both derivatives, gives the delta-method quantities for each $q$. The variance of a weighted mean uses squared weights. For $A=\bar q_a$, $B=\bar q_b$, and gain expressed in percent, the final variance is
\begin{equation}
\operatorname{Var}(G_q)\simeq10^4\left[
\frac{V_A}{B^2}+\frac{A^2V_B}{B^4}-\frac{2A C_{AB}}{B^3}
\right].
\end{equation}
Intervals are $G_q\pm1.96\sqrt{\operatorname{Var}(G_q)}$. The same formula with identity transform yields full-sequence failure intervals. These are approximate sampling intervals for fixed checkpoints, not intervals over retraining. The near-saturation derivative motivates retaining the untransformed full-grid analysis alongside the per-round analysis.

The supplied numerical archive stores the common-domain flag for every compound-noise condition, model-level error counts, and paired error summaries. Its validity mask is inherited from the original full evaluation panel and is not recomputed from the subset displayed in the manuscript. Recomputing per-round results on a different mask would change the estimand. The collection script independently checks paired-count identities, the 331-condition mask, the high-statistics architecture gain, AF/BWT, OOD gains, and all three Willow-length comparisons against the archived summaries.

\section{Architecture transfer across sequence lengths}
\label{app:lengths}
\begin{figure}[H]
\centering
\includegraphics[width=\textwidth]{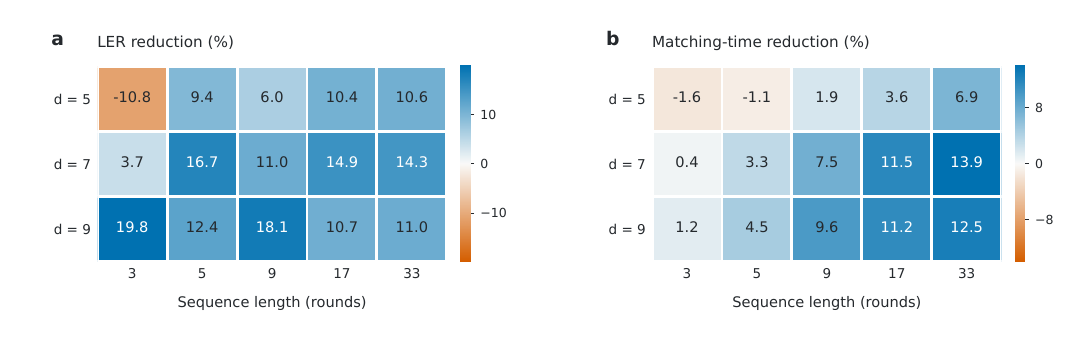}
\caption{\textbf{Complete architecture length extension.} HTNet relative to Ising-Fast, using fixed T0 base checkpoints. Positive values indicate reductions in per-round LER (left) or matching-backend time (right). Each condition contains 10,000 shots per basis. Cells are point estimates; paired LER intervals appear in Table~\ref{tab:archlength}.}
\label{fig:archlength}
\end{figure}

Figure~\ref{fig:archlength} and Table~\ref{tab:archlength} report all 15 combinations of $d\in\{5,7,9\}$ and $r\in\{3,5,9,17,33\}$ in the architecture length extension. They use the same base checkpoints as the main architecture comparison. Each point has 10,000 shots per basis, so the nine-round estimates in this extension are distinct from both the 65,536-shot comparison and the million-shot distance-nine evaluation. At short sequences some gains reverse; at 33 rounds, all three distances improve in both LER and backend time.

\begin{table}[H]
\centering\small
\caption{\textbf{Architecture length-extension values.} Backend reductions refer only to residual matching.}
\label{tab:archlength}
\input{data/architecture_lengths_table.tex}
\end{table}

\section{Compound-noise transfer across sequence lengths}
\label{app:oodlength}
\begin{figure}[H]
\centering
\includegraphics[width=\textwidth]{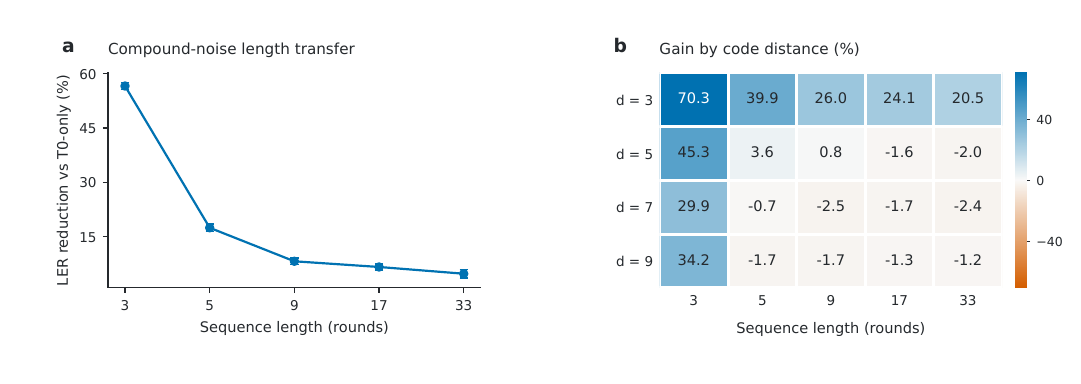}
\caption{\textbf{Length transfer under fixed compound noise.} (a) Distance-balanced EWC LER reduction relative to T0-only with paired 95\% intervals. (b) The same comparison resolved by distance. Each distance/length includes three noise combinations and 10,000 shots per basis and combination. The heatmap uses one symmetric color scale; values are relative percent reductions.}
\label{fig:oodlength}
\end{figure}

This experiment fixes the three noise combinations M+I, C+Z, and M+C+I+Z at multiplier $1.5$. The four distances and five lengths form a separate 60-condition grid. All per-round comparisons use the fixed domain common to the compared methods at each length. Figure~\ref{fig:oodlength} reports EWC relative to T0-only, using the final nine-round-trained weights. The large short-sequence gain and the positive long-sequence average coexist with nonpositive gains at larger distances for 17 and 33 rounds. This pattern differs from the fifty-round Willow result, where all three distance-level gains are positive.

\section{Willow length and distance decomposition}
\label{app:willow}
Table~\ref{tab:willowlength} lists absolute rates and paired gains for each native length and distance. The same patch/basis case identities are used at every length, but each length is evaluated on its own records. No pairing is assumed between different lengths. Distance-balanced means give equal weight to the three distances, rather than pooling all 28 cases as if their multiplicities were equal evidence about code-size dependence.

\begin{table}[H]
\centering\small
\caption{\textbf{Native Willow transfer by length and distance.} T0-only and EWC use HTNet. ``Mean'' weights distances equally. Wins count cases with lower point-estimate LER, not cases with individually significant differences.}
\label{tab:willowlength}
\input{data/willow_lengths_table.tex}
\end{table}

\section{Compound-noise families and the sequential control}
\label{app:ooddetail}
The family decomposition compares EWC and SEQ on the same valid conditions within each compound-noise family. The SEQ heatmap uses the same signed color scale and per-cell validity mask as the EWC heatmap in Fig.~\ref{fig:ood}c, so differences across strengths can be compared directly.

\begin{figure}[H]
\centering
\includegraphics[width=\textwidth]{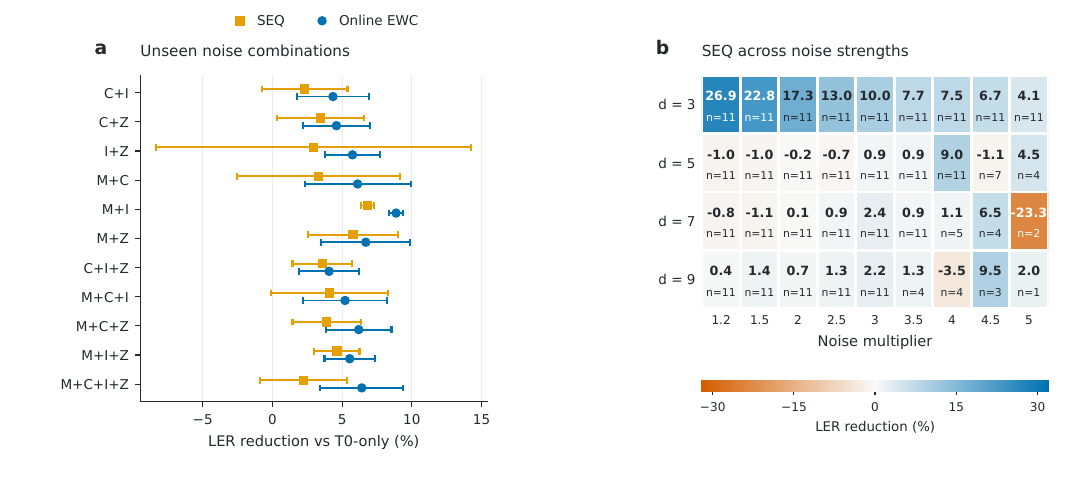}
\caption{\textbf{Noise-family and strength decomposition of frozen transfer.} (a) Per-round LER reduction relative to T0-only for all 11 compound-noise families, with paired 95\% intervals. M, C, I, and Z denote measurement, CNOT, idle, and Z-bias changes. (b) SEQ reduction by distance and multiplier, with the valid-condition count $n$ in each cell. Blue indicates improvement and orange deterioration. The fixed 331-condition validity mask, 10,000 shots per basis, and aggregation convention match Fig.~\ref{fig:ood}.}
\label{fig:ooddetail}
\end{figure}

\section{Architecture factors and transfer controls}
\label{app:additionalcontrols}
\begin{figure}[H]
\centering
\includegraphics[width=\textwidth]{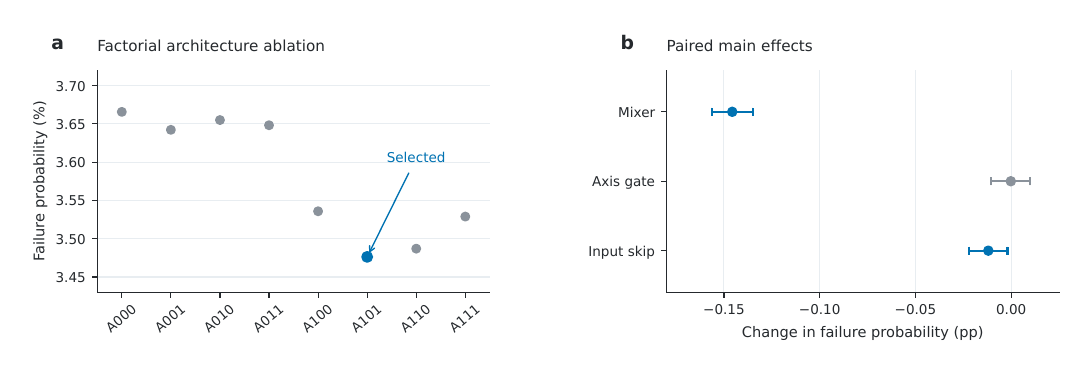}
\caption{\textbf{Complete factorial architecture evidence.} (a) Nine-round failure probability for all eight HTNet variants, averaging both bases with 1,048,576 shots per basis at $d=9$. Bits identify the post-fusion mixer, axis gate, and raw-input skip. A101 has the lowest observed mean. (b) Paired main effects in absolute percentage points, with approximate 95\% intervals; blue indicates significance after Holm correction across the three main effects. Adaptive branch fusion is shared by all variants.}
\label{fig:ablationdetail}
\end{figure}

\begin{figure}[H]
\centering
\includegraphics[width=\textwidth]{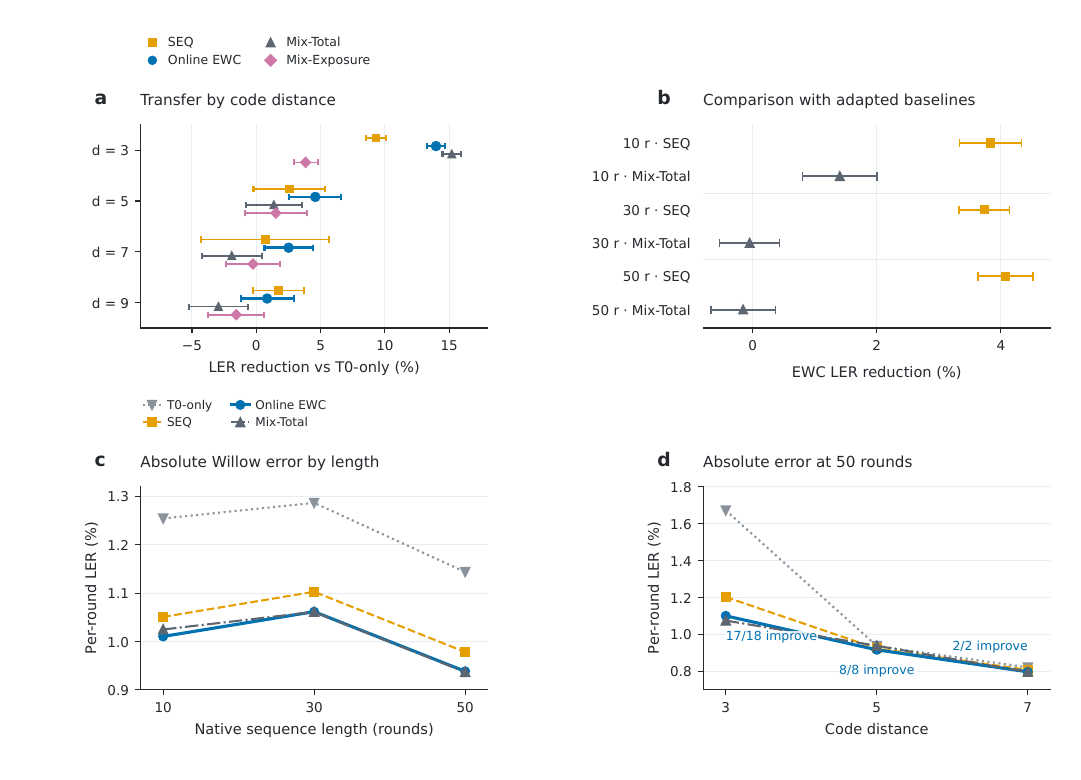}
\caption{\textbf{Distance-resolved and adapted-control comparisons.} (a) OOD per-round LER reductions relative to T0-only for SEQ, EWC, Mix-Total and Mix-Exposure, with paired 95\% intervals on the common validity domain. (b) Native Willow EWC reductions relative to SEQ and Mix-Total at each length, with paired 95\% intervals. (c) Absolute, distance-balanced per-round rates for the four neural training protocols across native lengths. (d) Absolute rates by distance at fifty rounds; annotations count EWC improvements over T0-only. Case identities, shot counts, and weighting match Fig.~\ref{fig:willow}.}
\label{fig:transfercontrols}
\end{figure}

Figure~\ref{fig:ablationdetail} retains the complete factorial comparison underlying the A101 selection. Figure~\ref{fig:transfercontrols} resolves transfer by distance and compares the adapted neural controls. EWC has a positive paired advantage over SEQ at each native Willow length, while its differences from Mix-Total span zero at 30 and 50 rounds. Absolute rates are retained alongside these relative comparisons.

\clearpage
\section{Input syndrome density across compound noise}
\label{app:inputdensity}
\begin{figure}[H]
\centering
\includegraphics[width=\textwidth]{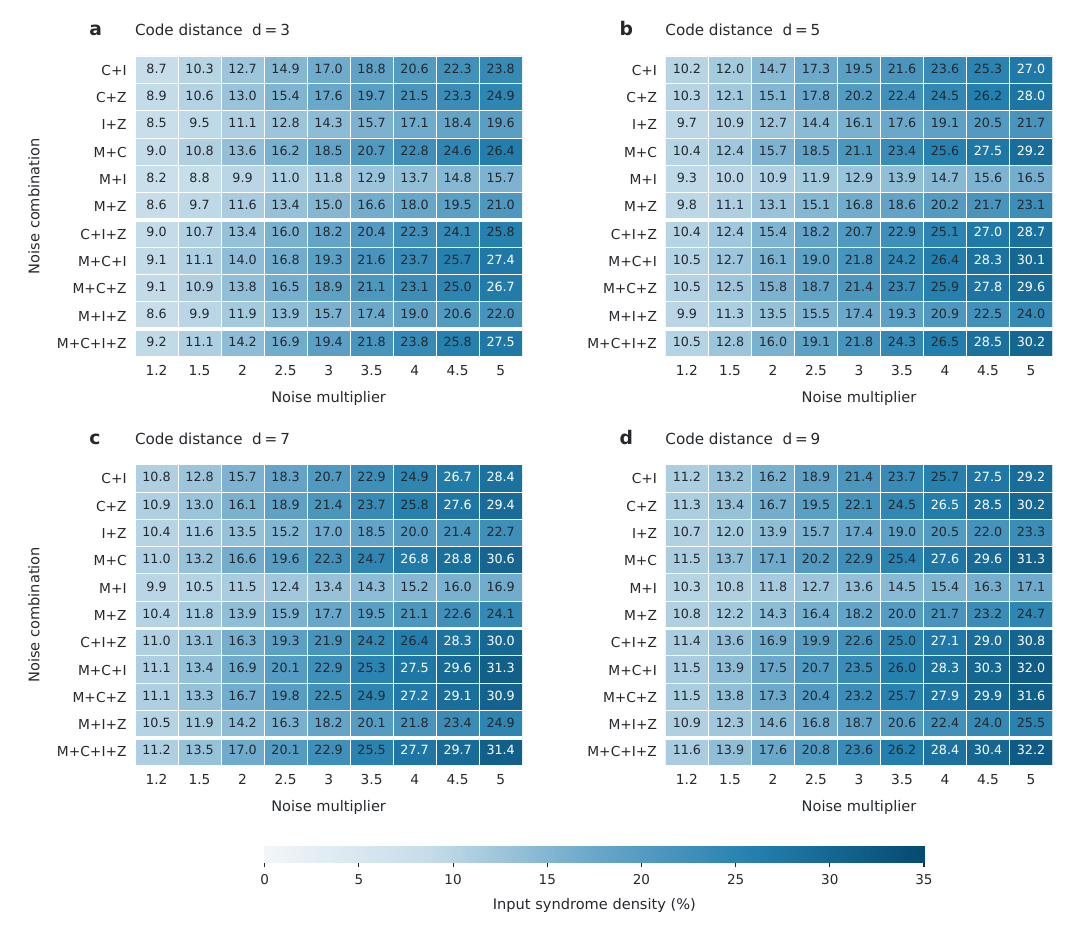}
\caption{\textbf{Input syndrome density across noise combinations and strengths.} (a--d) Code distances $3$, $5$, $7$, and $9$ at nine rounds. Rows show the 11 compound-noise combinations and columns the nine multipliers. Each cell gives the mean input density (\%), averaged over the X and Z logical bases, with 10,000 shots per basis. All 396 conditions are included. The four panels share a linear color scale from 0 to 35\%. M, C, I, and Z denote measurement, CNOT, idle, and Z-bias parameter changes.}
\label{fig:inputdensity}
\end{figure}

Figure~\ref{fig:inputdensity} maps the input syndrome density over the complete compound-noise grid. Density is the fraction of active detector entries before pre-decoding, computed within each logical basis and then averaged over X and Z. At $d=9$, it spans 10.26\%--32.17\%. At a fixed $5\times$ multiplier, M+I produces 17.12\% density, compared with 32.17\% for M+C+I+Z. Noise composition therefore changes the input detector load even at the same nominal multiplier.

\FloatBarrier
\end{document}

%% file: data/terminal_table.tex
\begin{tabular}{lrrrrr}
\toprule
Training method & ID LER (\%) & AF (pp) & BWT (pp) & OOD LER (\%) & Willow LER (\%)\\
\midrule
T0-only & 1.0766 & --- & --- & 6.0848 & 1.2541 \\
SEQ & 1.0715 & 0.0455 & -0.0432 & 5.8620 & 1.0507 \\
Online EWC & 1.0871 & 0.0107 & -0.0094 & 5.7404 & 1.0104 \\
Replay & 1.0966 & 0.0576 & -0.0576 & 6.4010 & 1.1602 \\
Mix-Total & 1.1757 & --- & --- & 5.8924 & 1.0248 \\
Mix-Exposure & 1.1424 & --- & --- & 6.0248 & 1.0814 \\
\bottomrule
\end{tabular}

%% file: data/ablation_table.tex
\begin{tabular}{lcccrr}
\toprule
Variant & $M$ & $A$ & $S$ & $p$ (\%) & $q$ (\%)\\
\midrule
A000 & 0 & 0 & 0 & 3.6655 & 0.4212 \\
A001 & 0 & 0 & 1 & 3.6421 & 0.4184 \\
A010 & 0 & 1 & 0 & 3.6549 & 0.4199 \\
A011 & 0 & 1 & 1 & 3.6481 & 0.4191 \\
A100 & 1 & 0 & 0 & 3.5358 & 0.4058 \\
A101 & 1 & 0 & 1 & 3.4762 & 0.3987 \\
A110 & 1 & 1 & 0 & 3.4870 & 0.4000 \\
A111 & 1 & 1 & 1 & 3.5289 & 0.4050 \\
\bottomrule
\end{tabular}

%% file: data/architecture_lengths_table.tex
\begin{tabular}{ccrrrr}
\toprule
$d$ & $r$ & Ising-Fast $q$ (\%) & HTNet $q$ (\%) & LER reduction, 95\% CI (\%) & Backend reduction (\%)\\
\midrule
5 & 3 & 0.4711 & 0.5221 & -10.83 [-20.54, -1.12] & -1.57 \\
5 & 5 & 0.6350 & 0.5751 & 9.43 [3.65, 15.21] & -1.06 \\
5 & 9 & 0.7083 & 0.6655 & 6.05 [1.71, 10.39] & 1.93 \\
5 & 17 & 0.7723 & 0.6922 & 10.37 [7.18, 13.56] & 3.62 \\
5 & 33 & 0.7923 & 0.7085 & 10.58 [7.73, 13.43] & 6.91 \\
7 & 3 & 0.3609 & 0.3474 & 3.74 [-7.21, 14.70] & 0.45 \\
7 & 5 & 0.4958 & 0.4128 & 16.74 [9.95, 23.53] & 3.25 \\
7 & 9 & 0.6150 & 0.5473 & 11.01 [6.15, 15.88] & 7.47 \\
7 & 17 & 0.6615 & 0.5630 & 14.90 [11.30, 18.50] & 11.50 \\
7 & 33 & 0.6837 & 0.5858 & 14.32 [11.36, 17.29] & 13.94 \\
9 & 3 & 0.2378 & 0.1907 & 19.79 [6.78, 32.81] & 1.21 \\
9 & 5 & 0.3632 & 0.3180 & 12.45 [3.86, 21.04] & 4.53 \\
9 & 9 & 0.4492 & 0.3679 & 18.10 [12.72, 23.48] & 9.63 \\
9 & 17 & 0.5179 & 0.4624 & 10.71 [6.45, 14.98] & 11.21 \\
9 & 33 & 0.5867 & 0.5222 & 10.99 [7.62, 14.37] & 12.49 \\
\bottomrule
\end{tabular}

%% file: data/willow_lengths_table.tex
\begin{tabular}{ccrrrr}
\toprule
Rounds & Distance & T0-only $q$ (\%) & EWC $q$ (\%) & Reduction, 95\% CI (\%) & Wins / cases\\
\midrule
10 & 3 & 1.8193 & 1.1146 & 38.73 [38.36, 39.11] & 17/18 \\
10 & 5 & 1.0405 & 1.0371 & 0.33 [-0.33, 0.98] & 5/8 \\
10 & 7 & 0.9026 & 0.8794 & 2.57 [1.11, 4.02] & 2/2 \\
10 & Mean & 1.2541 & 1.0104 & 19.44 [19.01, 19.86] & 24/28 \\
30 & 3 & 1.8417 & 1.2260 & 33.43 [33.01, 33.86] & 17/18 \\
30 & 5 & 1.0879 & 1.0634 & 2.25 [1.71, 2.79] & 8/8 \\
30 & 7 & 0.9295 & 0.8953 & 3.68 [2.52, 4.83] & 2/2 \\
30 & Mean & 1.2864 & 1.0616 & 17.47 [17.09, 17.86] & 27/28 \\
50 & 3 & 1.6713 & 1.1006 & 34.15 [33.50, 34.79] & 17/18 \\
50 & 5 & 0.9378 & 0.9168 & 2.24 [1.65, 2.84] & 8/8 \\
50 & 7 & 0.8219 & 0.7977 & 2.94 [1.73, 4.15] & 2/2 \\
50 & Mean & 1.1437 & 0.9384 & 17.95 [17.46, 18.44] & 27/28 \\
\bottomrule
\end{tabular}